\documentclass{article}
\usepackage{spconf}
\usepackage{hyperref}
\usepackage{url}
\usepackage{amsmath,amssymb,amsthm}
\usepackage{color}
\usepackage{graphicx}
\usepackage{subcaption}
\usepackage{floatrow}
\newfloatcommand{capbtabbox}{table}[][\FBwidth]
\usepackage{blindtext}
\usepackage{psfrag}
\usepackage{epstopdf}
\usepackage{float}
\usepackage{courier}

\newcommand{\bbE}{\ensuremath{\mathbb{E}}}

{%
\begin{list}{#1}{
\vspace{-\topsep}
\vspace{-\partopsep}
\setlength{\itemindent}{0cm}
\setlength{\rightmargin}{0cm}
\setlength{\listparindent}{0cm}
\settowidth{\labelwidth}{#1}
\setlength{\leftmargin}{\labelwidth}
\addtolength{\leftmargin}{\labelsep}
\setlength{\itemsep}{0cm}
}%
}%
{%
\end{list}
\vspace{-\topsep}
\vspace{-\partopsep}
}

{\begin{enumerate}%
}%
{\end{enumerate}}

\theoremstyle{plain}% default

\newtheorem*{prop*}{Proposition}

\theoremstyle{definition}

\newtheorem*{defn*}{Definition}

\newtheorem*{exmp*}{Example}

\newtheorem*{conj*}{Conjecture}

\theoremstyle{remark}

\newtheorem*{rmk*}{Remark}

\title{Acoustic feature learning \\ using
    cross-domain articulatory measurements}
\name{Qingming Tang$^*$, Weiran Wang$^{\dag 1}$\thanks{$^1$This work was completed while the author was working at TTIC.}
, Karen Livescu$^*$}
\vspace{-.05in}
\address{
  $^*$Toyota Technological Institute at Chicago, USA, $^\dag$Amazon Alexa, USA \\
  \texttt{$\{$qmtang,klivescu$\}$@ttic.edu, weiranw@amazon.com}}  
\vspace{-.05in}

\begin{document}
\ninept
\maketitle
\begin{abstract}
Previous work has shown that it is possible to improve speech recognition by learning acoustic features from paired acoustic-articulatory data, for example by using canonical correlation analysis (CCA) or its deep extensions.  One limitation of this prior work is that the learned feature models
are difficult to port to new datasets or domains, and articulatory data is not available for most speech corpora.
In this work we study the problem of acoustic feature learning in the setting where we have access to an external, domain-mismatched dataset of paired speech and articulatory measurements, either with or without labels.  We develop methods for acoustic feature learning in these settings, based on deep variational CCA and extensions that use both source and target domain data and labels. Using this approach, we improve phonetic recognition accuracies on both TIMIT and Wall Street Journal and analyze a number of design choices.

\end{abstract}

\vspace{-.07in}
\section{Introduction}
\label{sec:intro}
\vspace{-.05in}

Speech recognizers are typically trained on acoustic recordings along with their transcriptions.

In some cases we have access to additional data with another modality, such as video or articulation, and it may be fruitful to use this data as well for improved recognition.
Even if the additional modality is not available at test time, it may be possible to use,
e.g., to learn better acoustic features. For articulatory data in particular, it is possible to improve phonetic recognition
via multi-view feature learning using simultaneously recorded acoustic and articulatory data separate from the recognizer training data~\cite{arora2013multi,wang2015unsupervised,badino2016integrating,tang2017acoustic}.
These improvements hold for unseen speakers.
However, it is much more challenging to transfer this benefit to a new dataset from a different domain, where the recording conditions and linguistic material differ~\cite{arora2013multi}.

In this work we study how to transfer the potentially useful information that exists in an acoustic-articulatory dataset (the {\it source domain}, in this case the U.~Wisconsin X-ray Microbeam Dataset (XRMB)~\cite{westbury1990x}) to a recognizer in a different {\it target domain} (here, TIMIT~\cite{zue1990speech} or Wall Street Journal (WSJ)~\cite{paul1992design}).  We start from a successful recent approach for multi-view feature learning, deep variational canonical correlation analysis with private variables (VCCAP, Section~\ref{subsec:vcca})~\cite{wang2016deep,tang2017acoustic}, and consider ways of using it across domains. 

We investigate this problem in three settings, each with a different level of supervision: no labels for either the acoustic-articulatory source domain or the target recognizer domain during feature learning; target-domain recognition labels available in addition to unlabeled source-domain acoustic-articulatory data; and labels for both target dataset and source (acoustic-articulatory) dataset available during feature and recognizer training time.  We develop models for jointly training on data from both domains, which improve phonetic recognition performance over competitive baselines.

\vspace{-.05in}
\section{Related Work}
\label{sec:related}
\vspace{-.07in}

Multi-view feature (representation) learning has been studied for a variety of applications, typically using canonical correlation analysis (CCA)~\cite{hotelling1936relations, faruqui2014improving, yan2015deep}, contrastive losses~\cite{hermann2014multilingual, harwath2016unsupervised, he2016multi}, or other joint neural models~\cite{ngiam2011multimodal}, including acoustic feature learning with paired articulation~\cite{arora2013multi, wang2015unsupervised, tang2017acoustic, badino2016integrating}.
Other ways of using acoustic-articulatory measurement data, for example via articulatory inversion, have also been studied extensively~\cite{rose1996potential, frankel2001asr, ghosh2011automatic, mitra2017hybrid}.  Recent work has found that multi-view feature learning approaches that do not explicitly predict articulatory measurements tend to outperform articulatory inversion~\cite{wang2015unsupervised}, and that VCCAP outperforms other approaches~\cite{tang2017acoustic}, so this forms our starting point.
The setting where the multi-view data is also labeled has been studied less extensively; recent work found that a supervised extension of linear CCA can work well~\cite{arora2014multi}; here we explore this setting but with more powerful nonlinear models.

Our goal can be viewed as combining multi-view feature learning and adaptation to the target domain, so domain adaptation research is also relevant~\cite{ganin2016domain, tzeng2017adversarial, vincent2017analysis}, and speaker adaptation methods can be viewed as an instance of this~\cite{yao2012adaptation, karanasou2014adaptation, swietojanski2014learning, huang2016unified}.  It is possible to combine these adaptation techniques with multi-view feature learning, and we consider one such method in Section~\ref{subsec:extra_layers}; many more adaptation techniques can in principle be applied.

\vspace{-.07in}
\section{Cross-Domain Multi-View Feature Learning:  Unsupervised methods}
\label{sec:unsupervised}
\vspace{-.07in}

\begin{figure*}[t]
  \centering
    \includegraphics[width=0.76\linewidth]{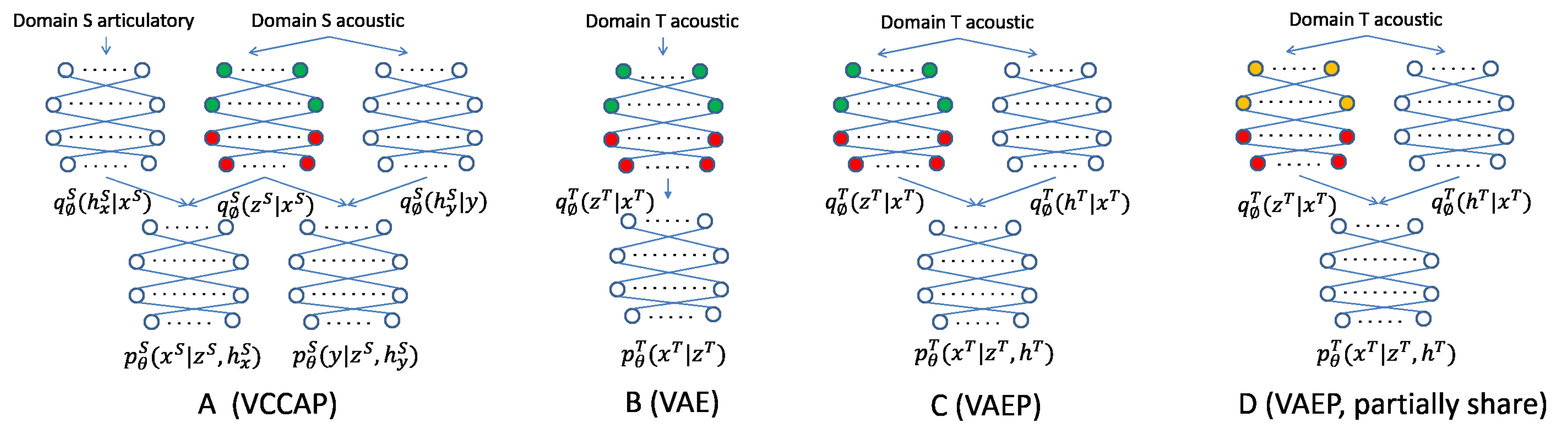}
  \vspace{-.1in}
  \caption{(A): VCCAP model for multi-view data; (B): Variational autoencoder (VAE) for the target-domain acoustics.  The projection network is the same as that of VCCAP (indicated by the colors); (C): Like (B), but with additional private variables for the target domain; (D): Like (C), but sharing only part of the projection network (in red) with VCCAP; the other layers (yellow) model domain-specific information.
}
  \label{fig:models}
\end{figure*}

In this section, we will consider the case where we learn acoustic features from the multi-view source dataset without accessing any labels for the source or target datasets. 

\vspace{-.1in}
\subsection{Variational deep CCA with private variables (VCCAP)}
\label{subsec:vcca}
\vspace{-.07in}

The basic multi-view model we begin with is {\it deep variational canonical correlation analysis with private variables} (VCCAP)~\cite{wang2016deep, tang2017acoustic}, shown in Fig.~\ref{fig:models}A.
VCCAP can be interpreted as a model for generating acoustic-articulatory data from latent variables that represent the information that is common to both views, combined with acoustic-specific and articulatory-specific latent variables that represent information that is ``private'' to one of the views. While the model appears in some ways quite complex,
it is easier to train and more successful in practice than earlier methods like deep CCA~\cite{wang2016deep}.

We use superscript $S$ to denote the source domain and $T$ for the target domain. $z^{S}$ in Fig.~\ref{fig:models}A represents information shared by the two views, which is hopefully discriminative information related to the phonetic labels.
However, there can be useful information that is not shared in the two views; thus two ``private'' variables $h_x$ and $h_y$ are introducted for representing acoustic- and articulatory-specific information respectively.  The acoustic measurements are assumed to be generated from $z^{S}$ and $h_x^S$, and the articulatory measurements from $z^S$ and $h_y^S$, via the reconstruction networks $p^{S}_{\theta}(\cdot)$.

The model uses variational inference~\cite{hoffman2013stochastic} and can be viewed loosely as a multi-view extension of variational autoencoders~\cite{kingma2013auto,rezende2014stochastic}. The projection network (colored green and red) outputs the mean and variance of a multidimensional Gaussian approximate posterior distribution $q_{\phi}^S(z^S|x^S)$. The learned acoustic features are simply this conditional Gaussian mean, which serves as an estimate of $z^S$; the rest of the network can be discarded after training.  We use only the shared variables $z^S$ and discard the view-specific ones $h^S$.\footnote{In initial experiments, using $h^S$ did not improve performance; this motivated the use of the target-domain private variables $h^T$ in the next section.} 
The objective function of VCCAP, given a single acoustic-articulatory training pair $(x^S,y)$, can be written as (see~\cite{wang2016deep})
\begin{eqnarray}
L_{VCCAP}(x^S,y) := -KL\big(q_{\phi}^S(z^S|x^S)||p(z^S)\big) \nonumber \\
-KL\big(q_{\phi}^S(h_x^S|x^S)||p(h_x^S)\big)-KL\big(q_{\phi}^S(h^S_y|y)||p(h^S_y)\big) \nonumber \\
+\bbE_{\{q_{\phi}^S(z^S|x^S)q_{\phi}^{S}(h_x^S|x^S)\}} \Big[ \log\big(p_{\theta}^S(x^S|z^S,h_x^S)\big)\Big] \nonumber \\
+\bbE_{\{q_{\phi}^S(z^S|x^S)q_{\phi}^{S}(h_y^S|y)\}} \Big[ \log\big(p_{\theta}^S(y|z^S,h_y)\big) \Big]
\label{eqn:vccap}
\end{eqnarray}
where $p(z^S)$, $p(h_y^S)$ and $p(h_x^S)$ are prior distributions of the latent variables, which are all $\mathcal{N}(0,I)$ here unless otherwise indicated.

\vspace{-.1in}
\subsection{Joint modeling of source and target domains}
\label{subsec:joint_model}
\vspace{-.08in}

Using the learned VCCAP network $q_\phi^S(z^S|x^S)$ directly in a target domain does not in general work well ``out of the box'' if there is significant domain mismatch.  Instead, we learn a projection network for the target domain, $q_\phi^T(z^T|x^T)$, that is informed by the source-domain model in various ways. One way is to have the two networks fully/partially share parameters, and train the two jointly in a unified model. Fig.~\ref{fig:models}B,C,D show several options for modeling the target-domain data. Architectures B, C, and D can each be combined with A to form three different models that can be viewed as ``weakly supervised'' by the cross-domain articulatory data. By ``combining'', here we mean that training is done with a loss that is a linear combination of the multiple relevant losses.

Model B represents the target-domain acoustics with a variational autoencoder (VAE), trained jointly with VCCAP with a shared projection network.
Model C (``VAEP'') is similar to B, but with an additional private variable $h^T$ and corresponding private projection network that is specific to the target domain.
Depending on the degree of domain mismatch, sharing the complete VCCAP network between source and target domains may still be too restrictive. Model D is similar to C, but with only a subset of the VCCAP layers shared.
The hidden layers that are closer to the acoustic input (in yellow) are treated as domain-specific, while the layers closer to the output features (in red) are shared between domains.
The objective function for C and D, for one acoustic frame $x^T$, can be written as:
\vspace{-.1in}
\begin{eqnarray}
L_{VAEP}(x^T) := \bbE_{\{q_{\phi}^T(z^T|x^T)q_{\phi}^T(h^T|x^T)\}} \Big[ \log\big(p_{\theta}^T(x^T|z^T,h^T) \big) \Big] \nonumber \\
-KL\big(q_{\phi}^T(z^T|x^T)||p(z^T)\big)-KL\big(q_{\phi}^T(h^T|x^T)||p(h^T)\big) 
\label{eqn:vaep}
\end{eqnarray} 
The objective for the combined model on $S$ and $T$ is
\vspace{-.05in}
\begin{equation}
(1-\beta)L_{VCCAP}(x^S,y^S) + \beta L_{VAEP}(x^T)
\label{eqn:unsupervised}
\end{equation}
where $\beta>0$ is a hyper-parameter and $p(h^T)$ and $p(z^T)$ are set to $\mathcal{N}(0,I)$. The feature vector used for downstream tasks is the mean of $q_{\phi}^T(z^T|x^T)$. 
In practice, we train all of the models with minibatch gradient descent methods. Using a joint loss for data from both domains is done by taking each minibatch to include some data drawn independently from each domain; for each domain-specific loss term we use the corresponding subset of the minibatch.

\vspace{-.05in}
\section{Supervised Approaches}
\label{sec:source+label}
\vspace{-.07in}

If target-domain labels are available, we may be able to do better than the unsupervised methods of the previous section.  For
concreteness, we use bidirectional long short-term memory (LSTM) recurrent neural networks (RNNs)~\cite{hochreiter1997long,schuster1997bidirectional} trained with the connectionist temporal classification (CTC) loss~\cite{graves2006connectionist}, which have recently achieved state-of-the-art results in ASR (e.g.,~\cite{zweig2017advances}).

\vspace{-.08in}
\subsection{Domain adaptation with extra layers}
\label{subsec:extra_layers}
\vspace{-.05in}

One way to use the learned features in a new domain is to add explicit domain adaptation layers (see Sec.~\ref{sec:related}).
In this approach, the projection network $q_{\phi}^S(z^S|x^S)$ is shared with the target domain.  However, two additional fully connected layers, one with ReLU~\cite{maas2013rectifier} activation and one linear transformation, is used to transform the target input data before it is fed to the VCCAP projection network.  The output of this composed projection network is the input to the recognizer. All training is done end-to-end. Such a simple model corresponds to ``VCCAP + adaptation layers'' in Table~\ref{tab:timit}. 

\vspace{-.08in}
\subsection{Joint training of target recognizer and features}
\label{subsec:joint_training}
\vspace{-.05in}
An alternative to explicit domain adaptation is to adapt implicitly, by keeping the feature projection structure fixed but jointly learning it along with the recognizer.  This may be preferable over adding extra layers, which can result in an overparameterized model.
To be more concrete, for the feature learning model we will use VCCAP+VAEP from the previous section, since (as will be shown in Sec.~\ref{sec:experiment}) it is the best-performing unsupervised model (although the approach in this section can be used with any of the feature learning losses).  

Denoting one target-domain acoustic utterance $\mathbf{x}^T$ and one frame $x^T$, the objective function of the multitask learning model (averaged over source and target datasets) is as follows:
\begin{eqnarray}
\alpha \big \{(1-\beta) L_{VCCAP}(x^{S},y) + \beta L_{VAEP}(x^T) \big\} \nonumber \\
+ (1-\alpha)  L_{CTC} ( \mathcal{F}_{VAEP}(\mathbf{x}^T) )
\label{eqn:multitask}
\end{eqnarray}
where $\mathcal{F}_{VAEP}(\mathbf{x}^T)$ is the sequence of means of $q_{\phi}^T(z^T|x_i^T)$ for all frames $i$ in $\mathbf{x}^T$; these are the learned features that are used as input to the target-domain recognizer. 
$\alpha$ is a tunable tradeoff parameter between the recognizer and feature learning losses.

\vspace{-.09in}
\subsection{Joint training of source and target recognizers}
\label{subsec:shared_blstm}
\vspace{-.07in}
Finally, if we have access to labels for both source and target domains, we may be able to benefit from jointly training recognizers for both domains, without direct use of the learned feature projection network in the target domain. 
In this approach, the source-domain recognizer uses VCCAP-based features fed into an LSTM-CTC recognizer, and the target-domain recognizer uses the original acoustic features fed into another LSTM-CTC recognizer.
We only share the topmost recurrent layer of the two recognizers for the two domains, which are trained jointly.
The idea here is to implicitly use the cross-domain articulatory data by encouraging the two recognizers to agree.  Although source-domain labels are present, the articulatory data may still help as a form of regularizer.
While this may seem like a very weak use of the articulatory data, this approach achieves 
good recognition improvements on the target domain (see Sec.~\ref{sec:experiment}).

\vspace{-.1in}
\section{Experiments}
\label{sec:experiment}
\vspace{-.05in}

We use three datasets: XRMB, TIMIT, and WSJ.
XRMB consists of $47$ speakers and $2357$ utterances.  
We use the standard TIMIT $3696$-utterance training set, $192$-utterance core test set, and a separate development set of $400$ utterances~\cite{povey2011kaldi}.  For WSJ, the (standard) training/dev/test sets consist of 37416/503/333 utterances.
The final task is phonetic recognition, evaluated using phonetic error rate (PER).
We consider three source-target domain pairs:

{\bf 1) XRMB(35) $\rightarrow$ XRMB(12)} This setup follows earlier work~\cite{wang2015unsupervised, wang2016deep, tang2017acoustic}.  We split the 47 XRMB speakers into two disjoint sets, consisting of 35 and 12 speakers respectively.  We treat the 35 speakers as the source ``domain'' and the 12 speakers as the target ``domain'', and we do not access the articulatory data for the target speakers.  We perform recognition experiments in a $6$-fold setup on the $12$ target speakers, where in each fold we train on $8$ speakers, tune on $2$, and test on $2$; we then report the average performance over the $6$ test sets.  This can be viewed as a very mild case of cross-domain learning.  As shown in prior work, in this setting we can improve target speaker performance by simply using features learned on the source speakers.  Our experiments in this setting are intended to ensure that our approaches still work in this mild case.

{\bf 2) XRMB $\rightarrow$ TIMIT} In this setting we use XRMB
as the source domain and TIMIT as the target domain.  One prior paper has explored an application of multi-view feature learning from XRMB to TIMIT, but in a more limited setting with fewer speakers and with shallow (kernel-based) feature learning models~\cite{arora2013multi}.

{\bf 3) XRMB $\rightarrow$ WSJ}  Here we use XRMB
as the source domain and WSJ as the target domain.  Whereas XRMB and TIMIT have similar amounts of data, WSJ is much larger, so we may expect that any external multi-view data will have a smaller effect.  We include both TIMIT and WSJ as target domains, both to test this possibility and more generally to measure applicability across target domains.

\vspace{.1in}
\noindent{\bf Experimental details.} 
 For XRMB/TIMIT, we use $39$-$D$ MFCCs as input. For WSJ we use $123$-$D$ log mel filterbank features ($40$-$D$ filter outputs plus energy, along with $1^\textrm{st}$ and $2^\textrm{nd}$ derivatives).  We use no speaker normalization/adaptation.  The XRMB input articulatory features are horizontal and vertical displacements of 8 pellets attached to the articulators ($16$-$D$ per frame).

We implement our models using TensorFlow~\cite{abadi2016tensorflow}. VCCAP models are trained on XRMB for $300$ epochs.  The RNN recognizers are $2$-layer bidirectional LSTMs trained for $50$ epochs for XRMB/WSJ and $80$ epochs for TIMIT; the epoch with the best dev PER is used for test set experiments. 
We optimize with Adam~\cite{kingma2014adam} for XRMB/WSJ; for TIMIT we tune the choice of vanilla SGD or Adam.  We use dropout~\cite{srivastava2014dropout} at a rate of $0.2-0.4$.  The batch size is $200$ frames for VCCAP and VAE(P); for recognizer training, the batch size is $1$/$2$/$16$ utterances for XRMB/TIMIT/WSJ.
We decode with beam search with beam size $100$/$200$/$50$ for XRMB/TIMIT/WSJ, using the algorithm of~\cite{maas2015lexicon}.
Hyperparameters ($\alpha$ (Eq.~\ref{eqn:multitask}), $\beta$ (Eq.~\ref{eqn:unsupervised} and ~\ref{eqn:multitask}), learning rate, dropout rate, tunable parameters for decoding, etc.) were tuned on the development sets. 

\vspace{-.15in}
\subsection{Main results}
\vspace{-.09in}

\begin{figure}[t]
\centering
\hspace{-.05in}\includegraphics[width=0.99\textwidth]{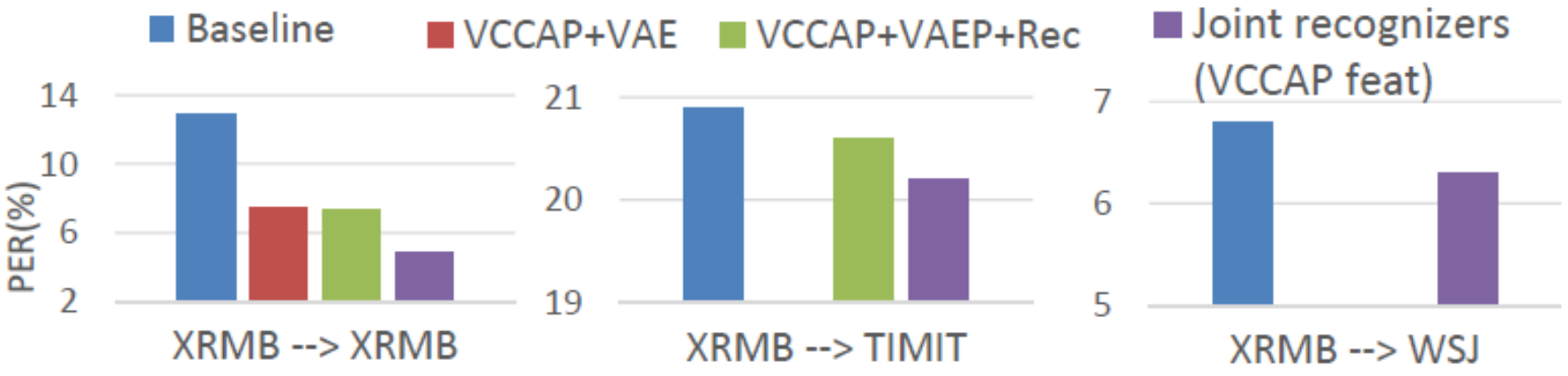}
\vspace{-.1in}
\caption{Summary of baselines and best final results.}
\label{fig:barplot}
\end{figure}

Fig.~\ref{fig:barplot} summarizes our baseline and best results. Detailed comparisons are given in Tables~\ref{tab:xrmb}--\ref{tab:wsj}, discussed in the following sections.

Our best model in the unsupervised XRMB(35)$\rightarrow$XRMB(12) setting is VCCAP+VAEP,
which reduces the PER from $12.9\%$ to $7.4\%$.  Using target-domain supervision, joint training with the RNN recognizer (Eq.~\ref{eqn:multitask}) gives a PER of $7.3\%$.
In the fully supervised setting, we obtain a PER of $4.9\%$ using jointly trained recognizers with VCCAP feature input.

In the XRMB $\rightarrow$ TIMIT and XRMB $\rightarrow$ WSJ settings, we experimented with a reduced set of models.  For XRMB $\rightarrow$ TIMIT, completely unsupervised methods fail.  Using target-domain labels, the best model is VCCAP+VAEP(partial)+Recognizer, similarly to the XRMB(35)$\rightarrow$XRMB(12) case but with only partial sharing of the VAEP projection to account for domain differences.  The best fully supervised approach, joint recognizers trained on VCCAP features, reduces PER from $20.9\%$ to $20.2\%$.  For WSJ, we experimented only with the fully supervised setting, where PER is improved from $6.8\%$ to $6.3\%$.

\vspace{-.12in}
\subsection{XRMB(35) $\rightarrow$ XRMB(12)}
\label{subsec:xrmb}
\vspace{-.05in}

\begin{table}[hbt]
  \caption{Detailed experiments for XRMB(35)$\rightarrow$XRMB(12):  PER (\%) averaged over 6 folds.}
  \label{tab:xrmb}
  \centering
  \begin{tabular}{|l||r|}
    \hline
 {\bf Method} & {\bf Test set PER} \\
	\hline\hline
 1. Baseline recognizer & 12.9 \\
 2. Baseline + windowing & 17.5 \\
        \hline
\multicolumn{2}{|l|}{\it Fully unsupervised} \\ \hline
 3. VCCAP & 9.4 \\
 4. VCCAP+VAEP &  {\bf 7.4} \\
 5. VAEP+VAEP & 14.2 \\
        \hline
\multicolumn{2}{|l|}{\it Unsupervised source domain, supervised target domain} \\ \hline        
 6. VCCAP+Recognizer & 8.9 \\
 7. VCCAP+VAEP+Recognizer & {\bf 7.3} \\
 8. VAEP+VAEP+Recognizer & 8.6 \\
  		\hline
\multicolumn{2}{|l|}{\it Supervised source + target domains} \\ \hline
 9. Joint recognizers (acoustic-only) & 5.9 \\
 10. Joint recognizers (VCCAP features) & {\bf 4.9} \\
 	    \hline
  \end{tabular}
\end{table}

Table~\ref{tab:xrmb} gives the XRMB test set results. Row $1$ is the baseline, i.e.~the RNN with MFCC inputs trained with CTC loss. Since our feature learning uses $15$-frame input windows, we also include (row $2$) a baseline RNN that uses windowed $15$-frame MFCCs, to confirm that any improvement is not due to windowing; in fact this baseline is much worse. Row $3$ uses acoustic features learned with VCCAP on XRMB(35), reproducing the setting of prior work~\cite{tang2017acoustic}.\footnote{The results here are better than those of~\cite{tang2017acoustic} due to improved optimization and tuning, and a new TensorFlow version.}  Row $4$ jointly learns the VCCAP projection on XRMB(35) and a VAEP projection on the target speakers XRMB(12), which produces the best unsupervised feature learning result.

Rows $6, 7$ are end-to-end versions of rows $3, 4$, which show the benefit of learning the features and recognizer jointly when the target-domain transcriptions are available at feature learning time.
Specifically, row $7$ corresponds to the multitask model that jointly learns VCCAP on the $35$-speaker ``source domain'' and VAEP and the RNN on the $8$-speaker ``target domain'' training set.  This model is best in the unsupervised source domain case.

One possibility is that the benefits come from the extra {\it acoustic} data. To check this, in rows $5, 8$ we replace the VCCAP projection with a VAEP applied to the source speakers' acoustics, trained jointly with the rest of the model as in rows $4, 7$.   Indeed, row $8$ also improves greatly over the baseline, but is still well behind our best multi-view approach.  In the fully unsupervised setting, the acoustic-only approach (row $5$) fails.  The large gap between rows $4$ and $5$, and between rows $7$ and $8$, indicates the advantage of using external acoustic-articulatory pairs over extra acoustics alone.

Finally, we consider the case where both ``domains'' are supervised, i.e.~we have transcriptions for all of XRMB.  Rows $9$, $10$ correspond to recognizers jointly trained on the $35$ source + $8$ target speakers, using only acoustic data vs.~using VCCAP features learned on the $35$-speaker multi-view data.\footnote{In this case we use VCCAP with a $71$-frame window acoustic input.}  Even in the fully supervised case, the multi-view approach still gives a $1\%$ improvement.

In these XRMB experiments, the source and target ``domains'' are very well matched, and we always use models with shared projection networks across domains.  In the next two subsections, we consider the two settings with much larger domain mismatch, and include experiments with partially shared projection networks.

\vspace{-.12in}
\subsection{XRMB $\rightarrow$ TIMIT}
\label{subsec:timit}
\vspace{-.07in}

\begin{table}[htb]
  \caption{PER (\%) for XRMB$\rightarrow$TIMIT. `Partial' = projection networks of the two domains are partially shared (Fig.~\ref{fig:models} A, D).}
  \label{tab:timit}
  \centering
  \begin{tabular}{|l||r|r|}
    \hline
{\bf Method} & {\bf Dev} & {\bf Test} \\
	\hline\hline

 1. Baseline & 19.2 & 20.9 \\
 2. Baseline + windowing & 22.4 & - \\
        \hline
\multicolumn{3}{|l|}{\it Fully unsupervised} \\ \hline
 3. VCCAP & 29.7 & - \\
 4. VCCAP+VAEP & 25.3 & - \\
 5. VCCAP+VAEP(partial) & 24.9 &  - \\
        \hline
\multicolumn{3}{|l|}{\it Unsupervised source domain, supervised target domain} \\ \hline
 6. VCCAP + adaptation layers & 19.0 & - \\
 7. VCCAP+VAEP+Recognizer  & 19.2 & - \\
 8. VCCAP+VAEP(partial)+Recognizer & {\bf 18.8} &  {\bf 20.6} \\
        \hline
\multicolumn{3}{|l|}{\it Supervised source + target domains} \\ \hline                
 9. Joint recognizers (acoustic input) & 18.8 & - \\
 10. XRMB+TIMIT recognizer (acoustic input) & 18.4 & - \\
 11. Joint recognizers +3 layers & {19.0} & - \\
 12. Joint recognizers (VCCAP features) & {\bf 18.1} & {\bf 20.2} \\
        \hline

  \end{tabular}
\end{table}

In Table~\ref{tab:timit}, row $1$ is the baseline RNN, and row $2$ again shows that windowing alone does not help. Row $3$ shows that directly using VCCAP learned on XRMB fails to generalize to TIMIT.
Row $4$ introduces domain-specific private variables; the improvement over row $3$ shows their benefit.  Row $5$ is similar to row $4$ but with a partially shared projection (Sec.~\ref{subsec:joint_model}).
Rows $6$, $7$ and $8$ use the target domain labels via end-to-end joint training of features and recognizer. Compared to the XRMB(35) $\rightarrow$ XRMB(12) case, we obtain a smaller improvement by learning features using XRMB, but there is still a good gain. 

Row $12$ corresponds to two domain-specific recognizers trained jointly with a final shared layer (Sec.~\ref{subsec:shared_blstm}), which produces our best results. Again, we check whether this improvement could be due solely to the extra acoustic data, by training a similar model on only the acoustic input; the result (row $9$) is worse, indicating that our improvements are not due to the extra acoustics alone.
%the cross-domain acoustic data only is less helpful than cross-domain acoustic-articulatory pairs.
Row $10$ corresponds to a single recognizer trained on the merged acoustic data of XRMB and TIMIT; this model does surprisingly well, but still worse than the best performer.  Row $11$ adds a $3$-layer DNN to row $9$, and takes as input $15$-frame concatenated MFCCs, to test whether any improvement may be due only to the additional structure of the VCCAP layers. This result is worse, verifying that the improvements are not solely due to the model structure.

\vspace{-.12in}
\subsection{XRMB $\rightarrow$ WSJ}
\label{subsec:wsj}
\vspace{-.07in}
Based on the success of the supervised joint recognizers approach in the XRMB $\rightarrow$ TIMIT setting, we only consider this approach for WSJ.
We again train source and target recognizers with the topmost layer shared, using either VCCAP features
or plain acoustic features
as input to the source-domain recognizer.  Again, using the XRMB articulatory data improves WSJ phonetic recognition, more so than the additional external acoustic data alone.

\begin{table}[htb]
    \caption{Phonetic error rates (\%) for XRMB$\rightarrow$WSJ experiments.}
  \label{tab:wsj}
  \centering
  \begin{tabular}{|l||r|r|}
    \hline
 {\bf Method} & {\bf Dev} & {\bf Test} \\
	\hline\hline

 1. Baseline & 8.3 & 6.8 \\
        \hline
 2. Joint recognizers (acoustic input) & 8.2 & 6.6 \\
        \hline
 3. Joint recognizers (VCCAP features) & {\bf 7.9} & {\bf 6.3} \\
        \hline 

  \end{tabular}
\end{table}

\vspace{-.15in}
\section{Conclusion}
\label{sec:conclusion}
\vspace{-.07in}
We have found that acoustic-articulatory data can be used to learn improved acoustic features for phonetic recognition, even when the multi-view data is from a different domain than the recognizer's data.  While it had been previously shown that improved acoustic features can be learned from acoustic-articulatory data, the cross-domain approach is much more practical.  We have also confirmed that the benefit does not come simply from having additional acoustic data, and that there is a benefit even when both the source and target domain data sets are labeled.  That is, the articulatory measurements provide a different kind of supervisory signal that is complementary to the acoustics and labels.  Further exploration is needed to compare VCCAP-based methods to other types of multi-view feature learning, and to study their applicability in word-level recognition.

\bibliographystyle{IEEEbib}
\bibliography{mybib}
\end{document}